\documentclass[]{article}
\usepackage{amssymb}
\usepackage{amsmath}
\usepackage{amsthm}
\usepackage{graphicx}
\usepackage{algorithm}
\usepackage{algorithmic}
\title{ SENNS: Sparse Extraction Neural NetworkS for Feature Extraction. }
\author{Abdulrahman Oladipupo Ibraheem \\ rahmanoladi@yahoo.com \\ \\ Computing and Intelligent Systems Research Group \\ Department of Computer Science and Engineering \\ Obafemi Awolowo University, Ile-Ife, Nigeria. }
\begin{document}
\maketitle
\begin{abstract}
The feature extraction problem occupies a central position in pattern recognition and machine learning. In this concept paper, drawing on ideas from optimisation theory, artificial neural networks (ANN), graph embeddings and sparse representations, I develop a novel technique, termed SENNS (Sparse Extraction Neural NetworkS), aimed at addressing the feature extraction problem. The proposed method uses (preferably deep) ANNs for projecting input attribute vectors to an output space wherein pairwise distances are maximized for vectors belonging to different classes, but minimized for those belonging to the same class, while simultaneously enforcing sparsity on the ANN outputs. The vectors that result from the projection can then be used as features in any classifier of choice. Mathematically, I formulate the proposed method as the minimisation of an objective function which can be interpreted, in the ANN output space, as a \textit{negative factor } of the sum of the squares of the pair-wise distances between output vectors belonging to different classes, added to a \textit{positive factor} of the sum of squares of the pair-wise distances between output vectors belonging to the same classes, plus sparsity and weight decay terms. To derive an algorithm for minimizing the objective function via gradient descent, I use the multi-variate version of the chain rule to obtain the partial derivatives of the function with respect to ANN weights and biases, and find that each of the required partial derivatives can be expressed as a sum of six terms. As it turns out, four of those six terms can be computed using the standard back propagation algorithm; the fifth can be computed via a slight modification of the standard backpropagation algorithm; while the sixth one can be computed via simple arithmetic. Finally, I propose experiments on the ARABASE Arabic corpora of digits and letters, the CMU PIE database of faces, the MNIST digits database, and other standard machine learning databases. 
\end{abstract}

\section{Introduction}
Most pattern recognition systems comprise three key stages: pre-processing, feature extraction and classification stages. Of these three stages, researchers believe that the feature extraction stage is the most critical. For example, in their review paper, the authors of \cite{trier} unequivocally wrote: \lq Selection of a feature extraction method is probably the single most important factor in achieving high recognition performance...\rq $\:$ Indeed, we agree with the authors of \cite{trier}, since our own view is that \textit{feature extraction is a commitment, once made might be irreversible by any classifier, however sophisticated}. Hence, it becomes highly paramount to carefully and rigorously study how these \lq commitments\rq $\:$ should be made --- how should features be extracted for optimal accuracies at the classification stage? While researchers have proposed a plethora of methods (e.g. \cite{eig_map}, \cite{Lda}, \cite{turk_pent}, \cite{hu}, \cite{ofmm}, \cite{sift}, \cite{shape_cont}, \cite{shock_graphs}, \cite{blum}, \cite{yan_emb}, \cite{ganesh} ) aimed at answering this question, it appears that a philosophy that should be followed by any good feature extraction method is the one articulated by Dejiver and Kittler in \cite{dejiver}. They said that feature extraction is the problem of \lq extracting from the raw data the information which is most relevant for classification purposes, in the sense of minimizing the within-class pattern variability, while enhancing the between-class pattern variability.\rq $\:$ Two of the more popular feature extraction techniques that follow this philosophy are the Linear Discriminant Analysis (LDA) \cite{Lda} and the Marginal Fisher Analysis (MFA) \cite{yan_emb}, both of which can be viewed as specific examples of a unifiying concept called graph embediings \cite{yan_emb}. Inspired by the above philosophy of \cite{dejiver}, we herein propose a technique, termed SENNS (Sparse Extraction Neural NetworkS) and pronounced \lq SENSE,\rq $\;$ for addreesing the feature extraction problem. Like the LDA and MFA, our SENNS can also be viewed, at least partially, via the lens of the graph embeddings concept. Unlike the MFA and LDA, however, SENNS incorporates a mechanism for seeking sparse features, and employs the apparatus of (preferably deep ) non-linear artificial neural networks, rather than linear (or kernel, or tensor) projections, for effecting the transformations that result in the sought features. Mathematically, we formulate our method as the minimisation of an objective function. Via rigorous mathematical analysis, we then derive a gradient descent algorithm for minimizing our objective function. Fortunately, it turns out that our algorithm can be expressed in terms of the standard backpropagation procedure, except, as we shall see, for a little tweaking to accomodate $L_1$ norms. Finally, we plan to test SENNS on standard machine learning datasets such as the ARABASE Arabic corpora of digits and letters \cite{arabase}, the CMU PIE database of faces \cite{cmu_pie}, the MNIST digits database \cite{mnist}, and other standard machine learning databases. 

\section{Notation}
In describing neural networks, I almost entirely follow the notation of Prof. Andrew Ng as in \cite{Ng}. In Ng's notation, for supervised training mode, the neural network learns from a training data denoted $(x^{(i)}, y^{(i)})$, $i = 1, 2, ..., m$. The neural network proper consists of $n_l$ layers, $L_1, L_2, ..., L_l, ..., L_{n_l}$, and the number of neurons in the $l$-th layer is denoted $s_l$. A weight, denoted $W_{ij}^{(l)}$, connects the $j$-th neuron of layer $l$ with the $i$-th neuron of layer $l + 1$, while a bias, denoted $b_i^{(l)}$, emanates from layer $l$ and enters the $i$-th neuron of layer $l + 1$. The overall function of the $i$-th neuron in layer $l$ is to compute an activation denoted $a_i^{(l)}$, which is the result of passing the quantity $z_i^{(l)} = \sum_{j =1} ^{s_l} w_{ij}^{(l)} a_j^{(l)} + b_i^{(l)}$ through a transfer function such as the sigmoid or tanh function. As a matter of notational expedience, for each $j$, the definition $a_i^{(1)} = x^{(j)}$ is employed, which means that the ANN input is viewed as an \lq activation\rq for the first layer. Furthermore, $a^{(l)}$ is used to denote the column vector given by $a^{(l)} = (a_1^{(l)}, a_2^{(l)}, ..., a_{s_l}^{(l)})^T$; and a similar notation applies to $b^{(l)}$ and $z^{(l)}$ as well. Similarly, $W^{(l)}$ denotes the column vector obtained via an ordered concatenation of all the weights linking layer $l$ with layer $l + 1$ in the network. Finally, $W$ represents a concatenation of all the weights in the ANN, while $b$ represents a concatenation of all the biases. 

\noindent Herein, I will follow the above notation, except for the following modifications. Firstly, to avoid confusion, I will use the index, $t$, placed within square brackets, to label training data: $(x^{{\tiny [t]}}, y^{{\tiny [t]}})$ $t = 1,2, \dots, m$. Based on this, I will write $a_i^{(l)[t]}$ to denote the activation from the $i$-th neuron of the $l$-th layer when $x^{[t]}$ is applied as input to the ANN, and I will use $a^{(l)[t]}$ to the denote the column vector of activations associated with the $l$-th layer and $t$-th input vector: $a_i^{(l)[t]} = (a_1^{(l)[t]}, a_2^{(l)[t]}, ..., a_{s_l}^{(l)[t]})^T$. However, when the layer in question is clear from context, I will simply write $a_i^{[t]}$ instead of $a_i^{(l)[t]}$, and $a^{[t]}$ instead of $a^{(l)[t]}$, in order to achieve a less clumsy notation. Finally, in dealing with partial derivatives, I shall herein write $\dfrac{\partial f}{\partial a^{{\tiny [t]}}} \text{{\Huge $\vert$}}_{\hat{a}^{{\tiny [t]}}} $ as a shorthand for $\dfrac{\partial f}{\partial a^{{\tiny [t]}}} \text{{\Huge $\vert$}}_{a^{{\tiny [t]}} = \hat{a}^{{\tiny [t]}}}$. Likewise, we shall write $\dfrac{\partial f}{\partial a^{{\tiny [t]}}} \text{{\Huge $\vert$}}_{(\hat{a}^{{\tiny [t]}}, \hat{a}^{{\tiny [u]}})}$ instead of $\dfrac{\partial f}{\partial a^{{\tiny [t]}}} \text{{\Huge $\vert$}}_{(a^{{\tiny [t]}}, a^{{\tiny [u]}}) = (\hat{a}^{{\tiny [t]}}, \hat{a}^{{\tiny [u]}})}$

\section{Formulation of Objective Function} Towards formulating the required objective function, I will begin by introducing two \lq functions\rq, $\mathbb{C}$ and $\mathbb{D}$ defined over the Cartesian product of input vectors according to:
\begin{equation} 
\mathbb{C}( x^{{\tiny [t]}}, x^{{\tiny [u]}} ) \; = \; 
\begin{cases} 1 & \text{if $x^{{\tiny [t]}}$ and $x^{{\tiny [u]}}$ belong to the \textbf{same} class} \\ 0 & \text{otherwise}
\end{cases}
\end{equation}
and
\begin{equation} 
\mathbb{D}( x^{{\tiny [t]}}, x^{{\tiny [u]}} ) \; = \; 
\begin{cases} 1 & \text{if $x^{{\tiny [t]}}$ and $x^{{\tiny [u]}}$ belong to \textbf{different} classes} \\ 0 & \text{otherwise}
\end{cases}
\end{equation}
Further, if $\mathcal{X} = \{ x^{{\tiny [1]}}, x^{{\tiny [2]}}, ..., x^{{\tiny [m]}}\}$ denotes the set of all input vectors, and $\mathcal{X}_2$ is the set of all pairs of the form $(x^{{\tiny [t]}}, x^{{\tiny [u]}})$, $t \in \{1,2, ..., m\}$ and $u \in \{1,2, ..., m\}$, which can be drawn from $\mathcal{X}$, then we shall use $M_ {\mathbb{C}}(\mathcal{X})$ to denote the number of times that function $\mathbb{C}$ outputs $1$ when all the pairs in $\mathcal{X}_2$ are passed through it. A similar definition applies to $M_ {\mathbb{D}}(\mathcal{X})$ as well. When the training set is clear from context, we will simply write $M_ {\mathbb{C}}$ and $M_ {\mathbb{D}}$ instead of $M_ {\mathbb{C}}(\mathcal{X})$ and $M_ {\mathbb{D}}(\mathcal{X})$ respectively. Next, for some non-negative regularisation constants, $\lambda_1, \lambda_2 \in [0, 1]$, we define a function, $\mathbb{S}( x^{{\tiny [t]}}, x^{{\tiny [u]}} )$, as follows: 
\begin{equation} 
\mathbb{S}( x^{{\tiny [t]}}, x^{{\tiny [u]}} ) \; = \; 
\begin{cases} \dfrac{\lambda_1}{M_ {\mathbb{C}}} & \text{if $x^{{\tiny [t]}}$ and $x^{{\tiny [u]}}$ belong to the \textbf{same} class} \\ - \dfrac{\lambda_2}{M_ {\mathbb{D}}} & \text{otherwise}
\end{cases}
\end{equation}
The connection between function $\mathbb{S}$ and the graph embedding framework \cite{yan_emb} should at once be clear. Specifically, it should be clear that $\mathbb{S}$ plays the role of the weights on the edges of the graphs underlying graph embeddings. We see that, similar to the Marginal Fisher Analysis (MFA) described in \cite{yan_emb}, $\mathbb{S}$ connects data points belonging to the same class with a positive weight, but connects those belonging to different classes with a negative weight. However, as we shall soon see, the proposed mathod herein differs from that in \cite{yan_emb} in three key ways. First, herein, we effect our projections via ANNs, unlike \cite{yan_emb} who employed either linearisation, kernelisation or tensorisation. Second, herein, we impose a sparsity requirement on the sought features, thereby seeking to take advantage of the well known benefits of sparse features; see \cite{sparse_code} and \cite{lap_sparse} for instance. Indeed, we would like to see the effect of the sparsity term on the ability of our gradient descent algorithm to locate a global minimum for our non-convex objective function. Thirdly, the overall structure of our objective function herein is different from that in \cite{yan_emb}, since ours is a regularised sum of terms, whereas theirs is a quotient of terms. The above three things also distinguish the method proposed herein from Linear Discriminant Analysis (LDA) \cite{Lda}.
\noindent We can now spell out the objective function we wish to minimize as follows:
\begin{multline}
J(W, b) = \dfrac{1}{2}\sum_{t= 1}^{m} \sum_{u= 1}^{m}\mathbb{S}( x^{{\tiny [t]}}, x^{{\tiny [u]}} ) || a^{{\tiny [t]}}\: - \: a^{{\tiny [u]}}||^2 \; \; + \; \; \dfrac{\lambda_3}{m} \sum_{t= 1}^{m} || a^{{\tiny [t]}} || _1 \; + \; \dfrac{\lambda_4}{2} \sum_{l= 1}^{n_l} \sum_{i= 1}^{s_{{ l + 1}}} \sum_{j= 1}^{s_l} W_{ij}^{(l)2} 
\end{multline}
\noindent
To achieve a slightly less clumsy notation in the above equation, we have written the activations associated with the ANN's output layer as $a^{{\tiny [t]}}$ instead of $a^{(n_l){\tiny [t]}}$. We shall carry on this practice henceforth, unless otherwise stated. Further, $||.||_1$ denotes the $L_1$ norm. Also, notice that the above objective function implicitly incorporates the regularizers, $\lambda_1, \lambda_2$, via the inclusion of $\mathbb{S}( x^{{\tiny [t]}}, x^{{\tiny [u]}} )$. In addition, $\lambda_3$ and $\lambda_4$ are also regularizers such that $\lambda_3, \lambda_4 \in [0, 1]$ and $\lambda_1 + \lambda_2 + \lambda_3 + \lambda_4 = 1$. By simply substituting $\mathbb{S}( x^{{\tiny [t]}}, x^{{\tiny [u]}} )$ from Equation 3 into Equation 4, we have: 
\begin{multline}
J(W, b) = \dfrac{\lambda_1}{2M_ {\mathbb{C}}}\sum_{t= 1}^{m} \sum_{u= 1}^{m}\mathbb{C}( x^{{\tiny [t]}}, x^{{\tiny [u]}} ) || a^{{\tiny [t]}}\: - \: a^{{\tiny [u]}}||^2 \; \; - \; \; \dfrac{\lambda_2}{2M_ {\mathbb{D}}}\sum_{t= 1}^{m} \sum_{u= 1}^{m}\mathbb{D}( x^{{\tiny [t]}}, x^{{\tiny [u]}} ) || a^{{\tiny [t]}}\: - \: a^{{\tiny [u]}}||^2 \\ + \dfrac{\lambda_3}{m} \sum_{t= 1}^{m} || a^{{\tiny [t]}} || _1 \; + \; \dfrac{\lambda_4}{2} \sum_{l= 1}^{n_l} \sum_{i= 1}^{s_{{ l + 1}}} \sum_{j= 1}^{s_l} W_{ij}^{(l)2} 
\end{multline}
\noindent
The form in Equation 5 above highlights the \lq graph embedding\rq \: aspect of our formulation. The first term gives a measure, in the output space of the ANN, of how widely separated output vectors belonging to the same class are. Clearly, we wish to minimize this non-negative quantity. On the contrary, the second term, \textit{excluding its negative sign}, gives a measure, in the output space of the ANN, of how widely separated output vectors belonging to different classes are. We wish to maximize this \textbf{non}-negative measure, by minimizing the negative quantity that results when the negative sign is pre-fixed to it. The third term is a sparsity term by which we wish to make the extracted features sparse. The fourth term is a weight decay term which prevents the weights from becoming too large, and helps prevent overfitting. Finally, we see that the parameters, $\lambda_1, \lambda_2, \lambda_3, \lambda_4$, allow us to control the relative amount of significance that the objective function attaches to the four objectives it is trying to achieve. 

\noindent A question naturally arises pertaining to the computational feasibility of the sums appearing in Equation 5. In particular, as we shall see, the sums involving index variables $t$ and $m$ will carry over directly to the algorithm for minimizing the objective function in Equation 5. This means that the term involving $\mathbb{C}( x^{{\tiny [t]}}, x^{{\tiny [u]}} )$ (as well as the term involving $\mathbb{D}( x^{{\tiny [t]}}, x^{{\tiny [u]}} )$) in the objective function would require $O(m^2)$ time, which becomes undesirable as $m$ grows. To ameliorate this, we are going to propose two heuristics for the two sums, as follows. We first consider the case of the term involving $\mathbb{D}( x^{{\tiny [t]}}, x^{{\tiny [u]}} )$. Upfront, we point out that the heuristic leads to a maximisation, in the ANN output space, of the sum of the distances between each input vector from a given class, and its nearest neighbour from each of the other classes, thereby given rise to a \textit{maximisation of the minimum distance formulation}, reminiscent of support vector machines \cite{svm}. To proceed, let $\mathcal{T} = \{ x^{{\tiny [1]}}, x^{{\tiny [2]}}, \dots, x^{{\tiny [m]}} \}$, and let there be $N$ classes, denoted $\Omega_1, \Omega_2, \dots, \Omega_N$ in the classification problem for which we are extracting features. Now, for each $x^{{\tiny [t]}} \in \Omega_p $, we define a set, $\mathbb{D}_{x^{{\tiny [t]}}}$ containing $N - 1$ elements as follows: $\mathbb{D}_{x^{{\tiny [t]}}} = \{d_1, d_2, \dots d_{p-1}, d_{p+1}, \dots, d_{N} \}$, such that each $d_p$ belongs to class $\Omega_{p}$ and each $d_p$ is the \textit{nearest} member of $\Omega_p$ from $x^{{\tiny [t]}}$. Our heuristic is to replace the quantity $\sum_{t= 1}^{m} \sum_{u= 1}^{m}\mathbb{D}( x^{{\tiny [t]}}, x^{{\tiny [u]}} ) || a^{{\tiny [t]}}\: - \: a^{{\tiny [u]}}||^2$ by $ \sum \limits_{x^{{\tiny [t]}} \in \mathcal{T}} \sum \limits_{d_q \in \mathbb{D}_{x^{{\tiny [t]}}}}|| a^{{\tiny [t]}}\: - \: a^{{\tiny [q]}}||^2$ where $a^{{\tiny [q]}}$ denotes the ANN output-layer vector of activations associated with input vector $d_q \in \mathbb{D}_{x^{{\tiny [t]}}}$. It should be clear that this new sum is $O(\gamma m)$, where $\gamma = N -1$, compared to the original sum which is $O(m^2)$. Since most problems usually have $m >> N - 1$, we expect this heuristic to lead to significant gains in computational feasibility in most cases. 

\noindent However, for the case of the term involving $\mathbb{C}( x^{{\tiny [t]}}, x^{{\tiny [u]}} )$, we propose an heuristic that leads to a situation wherein the sum of the distances between each input vector and its $k$-farthest neighbours, all in the same class as the input vector, is minimised. Formally, for each $x^{{\tiny [t]}} \in \Omega_p $, we simply define the set $\mathbb{C}_{x^{{\tiny [t]}}} = \tilde{N}_k(x^{{\tiny [t]}})$, where as usual, $\tilde{N}_k(x^{{\tiny [t]}})$ is the set of the $k$ \textit{farthest} elements from ${x^{{\tiny [t]}}}$ in $\Omega_p$. In Equation 5, we then replace the quantity $\sum_{t= 1}^{m} \sum_{u= 1}^{m}\mathbb{C}( x^{{\tiny [t]}}, x^{{\tiny [u]}} ) || a^{{\tiny [t]}}\: - \: a^{{\tiny [u]}}||^2$ by $ \sum \limits_{x^{{\tiny [t]}} \in \mathcal{T}} \sum \limits_{d_q \in \mathbb{C}_{x^{{\tiny [t]}}}}|| a^{{\tiny [t]}}\: - \: a^{{\tiny [q]}}||^2$. Again, in this case, we see that, for most cases, the heuristic can lead to improved computational feasibility since the new sum is $O(km)$, and $k$ can be chosen to be far lesser than $m$. 

\noindent Since the above heuristics constitute just an example of a host of possible heuristics that can be applied to alleviate the computational feasibilty issue in Equation 5, we therefore think that it would be better for us to develop our technique for the general case formulated in Equation 5, especially considering the fact that it should be clear how to adapt the developed technique to any particular heuristic of interest. So now, let us go back to Equation 4 (from which Equation 5 derives). It is expedient to denote the first term in the equation by $ J_1(W,b)$, the second term by $ J_2(W,b)$, and the third term by $J_3(W)$. Hence, Equation 4 can be re-written in the form: $J(W, b) = J_1(W,b) \; + \; J_2(W,b) \;+\; J_3(W)$

\noindent We now consider how to minimize $J(W, b)$ via gradient descent. A key step is the computation of $\nabla_{W^{(l)}} J(W, b)$ and $\nabla_{b^{(l)}} J(W, b)$, and this distills to the computation of $\dfrac{\partial J(W, b)}{\partial W_{ij}^{(l)} }$ and $\dfrac{\partial J(W, b)}{\partial b_{i}^{(l)} }$, for all $l = 1, 2,..., n_l - 1$, for all $j = 1, 2, ... s_l$, and for all $i = 1,2, ..., s_{l+1}$. I shall illustrate my overall approach by showing how to compute $\dfrac{\partial J(W, b)}{\partial W_{ij}^{(l)} }$, since the computation of $\dfrac{\partial J(W, b)}{\partial b_{i}^{(l)} }$ is analogous. To this end, I begin with a rather trivial step and write: 
\begin{equation}
\dfrac{\partial J(W, b)}{\partial W_{ij}^{(l)} } = \dfrac{\partial J_1(W,b)}{\partial W_{ij}^{(l)} } \; +\; \dfrac{\partial J_2(W,b)}{\partial W_{ij}^{(l)} } \; +\; \dfrac{\partial J_3(W)}{\partial W_{ij}^{(l)} } 
\end{equation}
\noindent But, the third partial derivative, $ \dfrac{\partial J_3(W)}{\partial W_{ij}^{(l)} }$, is particularly straightforward to compute: $\dfrac{\partial J_3(W)}{\partial W_{ij}^{(l)} } = W_{ij}^{(l)}$. Plugging this into Equation 6, we readily obtain: 
\begin{equation}
\dfrac{\partial J(W, b)}{\partial W_{ij}^{(l)} } = \dfrac{\partial J_1(W,b)}{\partial W_{ij}^{(l)} } \; +\; \dfrac{\partial J_2(W,b)}{\partial W_{ij}^{(l)} } \; +\; W_{ij}^{(l)} 
\end{equation}
\noindent To proceed, let us introduce two definitions: $\hat{J}_1(W,b) = \dfrac{1}{2} || a^{{\tiny [t]}}\: - \: a^{{\tiny [u]}}||^2$, and $\hat{J}_2(W,b) = || a^{{\tiny [t]}} || _1$. By comparing Equation 4 with the expression $J(W,b) = J_1(W,b) + J_2(W,b) J_3(W)$, observe that the first definition above allows us to write: $ J_1(W,b) = \sum \limits_{t= 1}^{m} \sum \limits_{u= 1}^{m}\mathbb{S}( x^{{\tiny [t]}}, x^{{\tiny [u]}} ) \hat{J}_1(W,b)$, so that:
\begin{equation} \dfrac{\partial J_1(W,b) }{\partial W_{ij}^{(l)}} = \sum \limits_{t= 1}^{m} \sum \limits_{u= 1}^{m}\mathbb{S}( x^{{\tiny [t]}}, x^{{\tiny [u]}} )\dfrac{\partial \hat{J}_1(W,b)}{\partial W_{ij}^{(l)}}
\end{equation}
In the same vein, notice that the second definition above permits us to write: $J_2(W,b) = \dfrac{\lambda_3}{m} \sum \limits_{t= 1}^{m} \hat{J}_2(W,b) $, so that:
\begin{equation}
\dfrac{\partial J_2(W,b)}{\partial W_{ij}^{(l)}} = \dfrac{\lambda_3}{m} \sum \limits_{t= 1}^{m} \dfrac{\partial \hat{J}_2(W,b)}{\partial W_{ij}^{(l)}}
\end{equation}. 
\noindent In rounding off this section, we should point out that the expression for $\dfrac{\partial J(W, b)}{\partial W_{ij}^{(l)} }$ given in Equation 7 will be particularly useful in the gradient descent algorithm we shall be proposing in Section 5 for minimizing $J(W, b)$. The algorithm will first use Equations 8 and 9 to compute $\dfrac{\partial J_1(W, b)}{\partial W_{ij}^{(l)} }$ and $\dfrac{\partial J_2(W, b)}{\partial W_{ij}^{(l)} }$ respectively; it will then use Equation 7 to compute $\dfrac{\partial J(W, b)}{\partial W_{ij}^{(l)} }$, before proceeding to update ANN weights in a gradient-descent style. However, it is clear from Equations 8 and 9 that if the algorithm is to be effective, then we must find ways of computing $\dfrac{\partial \hat{J}_1(W, b)}{\partial W_{ij}^{(l)} }$ and $\dfrac{\partial \hat{J}_2(W, b)}{\partial W_{ij}^{(l)} }$. The next two sections are devoted to these two tasks.
\section{Deriving the Partial Derivatives of $\hat{J}_1(W,b)$, and an Algorithm for Computing them.}
In this section, we focus on the computation of $\dfrac{\partial \hat{J}_1(W,b)}{\partial W_{ij}^{(l)} }$. From the preceding section, we know: \\
\begin{equation}
\hat{J}_1(W,b) = \dfrac{1}{2} || a^{{\tiny [t]}}\: - \: a^{{\tiny [u]}}||^2
\end{equation}
\noindent Next, I define vector $A$ to be the column vector formed by stacking the column vector, $a^{{\tiny [t]}}$ atop the column vector $a^{{\tiny [u]}}$. That is, $A = (a^{{\tiny [t]}^T}, a^{{\tiny [u]}^T} )^T$, where $a^{{\tiny [t]}^T}$ denotes the transpose of $a^{{\tiny [t]}}$. Notice that $\hat{J}_1(W,b)$ is a function of $A$, so that one may speak of computing $\nabla_ {\!A} \hat{J}_1(W, b)$, $\nabla_{\! a^{\tiny [t]}} \hat{J}_1(W, b)$ and $\nabla_{\!a^{\tiny [u]}} \hat{J}_1(W, b)$. Also, observe that, since each of $a^{{\tiny [t]}}$ and $a^{{\tiny [u]}}$ is a function of ANN weights, $W$, and biases, $b$, it follows that $A$ is also a function of weights and biases, so that one can as well speak of computing $\dfrac{\partial A }{\partial W_{ij}^{(l)} }$. Now, going back to Equation 10, we can now use the chain rule to write an expression for $\dfrac{\partial \hat{J}_1(W, b)}{\partial W_{ij}^{(l)} }$: 
\begin{equation}
\dfrac{\partial \hat{J}_1(W, b)}{\partial W_{ij}^{(l)} } \;=\; \text{{\huge (}}\dfrac{\partial A }{\partial W_{ij}^{(l)} }\text{{\huge )}}^{\!T} \nabla_ {\!A} \hat{J}_1(W, b)
\end{equation}
\noindent Being a dot product of two vectors, the expression above is a sum of terms. We can break the sum into two parts, one associated with the column vector, $a^{{\tiny [t]}}$ and the other associated with the column vector $a^{{\tiny [u]}}$: 
\begin{equation}
\dfrac{\partial \hat{J}_1(W, b)}{\partial W_{ij}^{(l)} } \;=\; \text{{\huge (}}\dfrac{\partial a^{{\tiny [t]}}}{\partial W_{ij}^{(l)} }\text{{\huge )}}^{\!T}\nabla_ {\!a^{{\tiny [t]}}} \hat{J}_1(W, b) \; + \; \text{{\huge (}}\dfrac{\partial a^{{\tiny [u]}}}{\partial W_{ij}^{(l)} }\text{{\huge )}}^{\!T} \nabla_ {\!a^{{\tiny [u]}}} \hat{J}_1(W, b)
\end{equation}
\noindent I will now go down a path to argue that each of the terms in Equation 12 above can be computed using the standard backpropagation algorithm. To this end, I consider a function that has a very similar form to $\hat{J}_1(W, b)$, except for a very small difference. In particular, I consider the function $\tilde{B}(a^{{\tiny [t]}} ; \; c^{{\tiny [t]}} )$, which I herein term a \textit{Back-Propagatable} function, and which I define according to: 
\begin{equation}
\tilde{B}(a^{{\tiny [t]}} ; \; c^{{\tiny [t]}} ) = \dfrac{1}{2} || a^{{\tiny [t]}}\: - c^{{\tiny [t]}} ||^2
\end{equation}
where, $c^{{\tiny [t]}} $ is a constant that is independent of the ANN's weights and biases. We ask the reader to compare the function $\tilde{B}(a^{{\tiny [t]}} ; \; c^{{\tiny [t]}} ) = \dfrac{1}{2} || a^{{\tiny [t]}}\: - c^{{\tiny [t]}} ||^2$ with the function $\hat{J}_1(W, b) = \dfrac{1}{2}|| a^{{\tiny [t]}}\: - a^{{\tiny [u]}} ||^2$, defined in Equation 10. In particular, the reader should note that $\tilde{B}(a^{{\tiny [t]}} ; \; c^{{\tiny [t]}} )$ can be obtained from $E(W, b ; \; x^{{\tiny [t]}}, x^{{\tiny [u]}} )$ simply by replacing the \textit{function} (of weights and biases) $a^{{\tiny [u]}}$ in the latter by the \textit{constant} $c^{{\tiny [t]}}$. However, what is more important is that the back-propagatable function in Equation 13 above plays a central role in the expression of the sum of squares error that an ANN aimed at classification must try to minimize in supervised learning mode. Given training data, $\mathcal{T} = (x^{{\tiny [t]}}, y^{{\tiny [t]}}), t = 1,2,3,..., m$, I can recall that that sum of squares error can be written as: 
\begin{equation}
B(W, b) = \dfrac{1}{2m}\sum_{t =1}^m || a^{{\tiny [t]}}\: - y^{{\tiny [t]}} ||^2
\end{equation}
\noindent Clearly, in the above equation, $y^{{\tiny [t]}}$ is a constant independent of ANN weights and biases. Moreover, it is also clear that the expression inside the summation on the right hand side of the equation perfectly fits into:
\begin{equation}
\tilde{B}(a^{{\tiny [t]}} ; \; y^{{\tiny [t]}}) = \dfrac{1}{2}||a^{{\tiny [t]}}\: - y^{{\tiny [t]}} ||^2
\end{equation}
To proceed, we put Equation 15 into Equation 14 and obtain: 
\begin{equation}
B(W, b) = \dfrac{1}{m}\sum_{t =1}^m \tilde{B}(a^{{\tiny [t]}} ; \; y^{{\tiny [t]}})
\end{equation}
\noindent In supervised learning mode, the objective of the ANN is to minimize the total error, $B(W, b)$.One of the most frequently employed techniques for minimizing $B(W, b)$ is the gradient descent algorithm, and this requires the partial derivatives of $B(W, b)$ with respect to weights and biases. Now, the partial derivative of $B(W, b)$ with respect to an arbitrary weight $W^{(l)}_{ij}$ can be expressed as:
\begin{equation}
\dfrac{\partial B(W, b)} {\partial W^{(l)}_{ij}} = \dfrac{1}{m}\sum_{t =1}^m \dfrac{\partial \tilde{B}(a^{{\tiny [t]}} ; \; y^{{\tiny [t]}}) } {\partial W^{(l)}_{ij}}
\end{equation}
So in essence, the task of computing $\dfrac{\partial B(W, b)} {\partial W^{(l)}_{ij}}$ boils down to the computation of $\dfrac{\partial \tilde{B}(a^{{\tiny [t]}} ; \; y^{{\tiny [t]}}) } {\partial W^{(l)}_{ij}}$, and then summing over all $t \in \{1,2, ..., m\}$ (i.e. summing over the training data); \textit{{ the backpropagation algorithm is normally employed for the computation of }} $\dfrac{\partial \tilde{B}(a^{{\tiny [t]}} ; \; y^{{\tiny [t]}}) } {\partial W^{(l)}_{ij}}$. \textit{{Generalizing, we see that the backpropagation algorithm can always be used to compute the partial derivatives}}, $\dfrac{\partial \tilde{B}(a^{{\tiny [t]}} ; \; c^{{\tiny [t]}} ) } {\partial W^{(l)}_{ij}}$, \textit{\textbf{of any function of the form}} $\tilde{B}(a^{{\tiny [t]}} ; \; c^{{\tiny [t]}} ) = \dfrac{1}{2} || a^{{\tiny [t]}}\: - c^{{\tiny [t]}} ||^2$, \textit{{such that}} $c^{{\tiny [t]}}$ \textit{is a constant} which plays the role which the target output value, $y^{{\tiny [t]}}$, plays in the supervised learning mode of ANNs. To proceed from here, we shall let $\hat{a}^{{\tiny [u]}}$ denote the specific constant value that the function $a^{{\tiny [u]}}$ ($a^{{\tiny [u]}}$ is a function of ANN weights and biases) evaluates to for a given value of $u$ and a given set of weights and biases. We then consider the function $\tilde{B}(a^{{\tiny [t]}} ; \; \hat{a}^{{\tiny [u]}} )$, in which $\hat{a}^{{\tiny [u]}}$ is playing the role that the target output value, $y^{{\tiny [t]}}$, plays in the function $\tilde{B}(a^{{\tiny [t]}} ; \; y^{{\tiny [t]}} )$. With the foregoing in mind, let us now try to write an expression for $\dfrac{\partial \tilde{B}(a^{{\tiny [t]}} ; \; \hat{a}^{{\tiny [u]}} )} {\partial W^{(l)}_{ij}}$. With the aid of the chain rule, we have:
\begin{equation}
\dfrac{\partial \tilde{B}(a^{{\tiny [t]}} ; \; \hat{a}^{{\tiny [u]}} )} {\partial W^{(l)}_{ij}} \;=\; \text{{\huge (}}\dfrac{\partial a^{{\tiny [t]}}}{\partial W_{ij}^{(l)} }\text{{\huge )}}^{\!T} \nabla_ {\!a^{{\tiny [t]}}} \tilde{B}(a^{{\tiny [t]}} ; \; \hat{a}^{{\tiny [u]}} )
\end{equation}
We now try to compare the quantity, $\text{{\huge (}}\dfrac{\partial a^{{\tiny [t]}}}{\partial W_{ij}^{(l)} }\text{{\huge )}}^{\!T} \nabla_ {\!a^{{\tiny [t]}}} \tilde{B}(a^{{\tiny [t]}} ; \; \hat{a}^{{\tiny [u]}} )$ on the right hand side of Equation 18 above with the quantity, $\text{{\huge (}}\dfrac{\partial a^{{\tiny [t]}}}{\partial W_{ij}^{(l)} }\text{{\huge )}}^{\!T}\nabla_ {\!a^{{\tiny [t]}}} \hat{J}_1(W, b) $, which occurs as the first term on the right hand side of Equation 12. We claim that, for a given value, $\hat{a}^{{\tiny [u]}}$, both are equal, in the sense that: $ \text{{\Huge [}} \text{{\huge (}}\dfrac{\partial a^{{\tiny [t]}}}{\partial W_{ij}^{(l)} }\text{{\huge )}}^{\!T}\nabla_ {\!a^{{\tiny [t]}}} \hat{J}_1(W, b) \text{{\Huge ]}} \text{{\Huge $\vert$}} _{\hat{a}^{{\tiny [u]}}} \: = \: \text{{\huge (}}\dfrac{\partial a^{{\tiny [t]}}}{\partial W_{ij}^{(l)} }\text{{\huge )}}^{\!T} \nabla_ {\!a^{{\tiny [t]}}} \tilde{B}(a^{{\tiny [t]}} ; \; \hat{a}^{{\tiny [u]}} )$, where, as explained in the section on notation, we have written $\text{{\Huge [}} \text{{\huge (}}\dfrac{\partial a^{{\tiny [t]}}}{\partial W_{ij}^{(l)} }\text{{\huge )}}^{\!T}\nabla_ {\!a^{{\tiny [t]}}} \hat{J}_1(W, b) \text{{\Huge ]}} \text{{\Huge $\vert$}} _{\hat{a}^{{\tiny [u]}}}$ instead of $\text{{\Huge [}} \text{{\huge (}}\dfrac{\partial a^{{\tiny [t]}}}{\partial W_{ij}^{(l)} }\text{{\huge )}}^{\!T}\nabla_ {\!a^{{\tiny [t]}}} \hat{J}_1(W, b) \text{{\Huge ]}} \text{{\Huge $\vert$}}_{a^{{\tiny [u]}} = \hat{a}^{{\tiny [u]}}}$. Before we show the equality, let us first highlight what we stand to gain if they are truly equal. In particular, $ \text{{\Huge [}} \text{{\huge (}}\dfrac{\partial a^{{\tiny [t]}}}{\partial W_{ij}^{(l)} }\text{{\huge )}}^{\!T}\nabla_ {\!a^{{\tiny [t]}}} \hat{J}_1(W, b) \text{{\Huge ]}} \text{{\Huge $\vert$}} _{\hat{a}^{{\tiny [u]}}} \: = \: \text{{\huge (}}\dfrac{\partial a^{{\tiny [t]}}}{\partial W_{ij}^{(l)} }\text{{\huge )}}^{\!T} \nabla_ {\!a^{{\tiny [t]}}} \tilde{B}(a^{{\tiny [t]}} ; \; \hat{a}^{{\tiny [u]}} )$ clearly implies \\ $\text{{\Huge [}} \text{{\huge (}}\dfrac{\partial a^{{\tiny [t]}}}{\partial W_{ij}^{(l)} }\text{{\huge )}}^{\!T}\nabla_ {\!a^{{\tiny [t]}}} \hat{J}_1(W, b) \text{{\Huge ]}} \text{{\Huge $\vert$}} _{\hat{a}^{{\tiny [u]}}} \: = \: \dfrac{\partial \tilde{B}(a^{{\tiny [t]}} ; \; \hat{a}^{{\tiny [u]}} )} {\partial W^{(l)}_{ij}}$. But, according to our discussion in the previous paragraph, we know that $\dfrac{\partial \tilde{B}(a^{{\tiny [t]}} ; \; \hat{a}^{{\tiny [u]}} )} {\partial W^{(l)}_{ij}}$ can be computed via backpropagation. Thus, if the equality is really true, we would have in essence found a means of computing $\text{{\Huge [}} \text{{\huge (}}\dfrac{\partial a^{{\tiny [t]}}}{\partial W_{ij}^{(l)} }\text{{\huge )}}^{\!T}\nabla_ {\!a^{{\tiny [t]}}} \hat{J}_1(W, b) \text{{\Huge ]}} \text{{\Huge $\vert$}} _{\hat{a}^{{\tiny [u]}}}$. Now, this would be very important to us because, as given in Equation 12, $ \text{{\huge (}}\dfrac{\partial a^{{\tiny [t]}}}{\partial W_{ij}^{(l)} }\text{{\huge )}}^{\!T}\nabla_ {\!a^{{\tiny [t]}}} \hat{J}_1(W, b) $ is one of the two terms involved in our expression for $\dfrac{\partial \hat{J}_1(W, b)}{\partial W_{ij}^{(l)} }$, which in turn is a partial derivative which we need for minimizing our objective function via gradient descent. 
\noindent Indeed, it is \textit{very simple} to show the required equality. Specifically, to show \\ $ \text{{\Huge [}} \text{{\huge (}}\dfrac{\partial a^{{\tiny [t]}}}{\partial W_{ij}^{(l)} }\text{{\huge )}}^{\!T}\nabla_ {\!a^{{\tiny [t]}}} \hat{J}_1(W, b) \text{{\Huge ]}} \text{{\Huge $\vert$}} _{\hat{a}^{{\tiny [u]}}} \: = \: \text{{\huge (}}\dfrac{\partial a^{{\tiny [t]}}}{\partial W_{ij}^{(l)} }\text{{\huge )}}^{\!T} \nabla_ {\!a^{{\tiny [t]}}} \tilde{B}(a^{{\tiny [t]}} ; \; \hat{a}^{{\tiny [u]}} )$, all we need show is $\nabla_ {\!a^{{\tiny [t]}}} \hat{J}_1(W, b) \text{{\Large $\vert$}} _{\hat{a}^{{\tiny [u]}}} \:=\: \nabla_ {\!a^{{\tiny [t]}}} \tilde{B}(a^{{\tiny [t]}} ; \; \hat{a}^{{\tiny [u]}} )$, since $\text{{\huge (}}\dfrac{\partial a^{{\tiny [t]}}}{\partial W_{ij}^{(l)} }\text{{\huge )}}^{\!T} \text{{\Huge $\vert$}} _{\hat{a}^{{\tiny [u]}}} \:=\: \text{{\huge (}}\dfrac{\partial a^{{\tiny [t]}}}{\partial W_{ij}^{(l)} }\text{{\huge )}}^{\!T} $ . A simple \lq proof\rq \; follows thus: \\ \\ \\
\noindent \begin{proof} [\lq \textbf{Proof}\rq \: of \; $ \nabla_ {\!a^{{\tiny [t]}}} \hat{J}_1(W, b) \text{{\Large $\vert$}} _{\hat{a}^{{\tiny [u]}}} \:=\: \nabla_ {\!a^{{\tiny [t]}}} \tilde{B}(a^{{\tiny [t]}} ; \; \hat{a}^{{\tiny [u]}} )$ ] \mbox{} \\
\noindent We begin with $\hat{J}_1(W, b) = \dfrac{1}{2} || a^{{\tiny [t]}}\: - \: a^{{\tiny [u]}}||^2 = \dfrac{1}{2} \sum_{i = 1}^{s_{n_l}} (a_i^{{\tiny [t]}}\: - \: a_i^{{\tiny [u]}})^2$, where $a_i^{{\tiny [t]}}$ denotes the activation of the $i$-th neuron in the output layer when the training example $x^{{\tiny [t]}}$ is applied as input to the ANN, and $s_{n_l}$ is the number of neurons in the output layer. We then find $\nabla_ {\!a^{{\tiny [t]}}} \hat{J}_1(W, b) = (a_1^{{\tiny [t]}} - a_1^{{\tiny [u]}}, \; a_2^{{\tiny [t]}} - a_2^{{\tiny [u]}}, \; \dots \: , \; a_{s_{n_l}}^{{\tiny [t]}} - a_{s_{n_l}}^{{\tiny [u]}} )$. Hence, for a given output value $\hat{a}^{{\tiny [u]}}$ of function ${a}^{{\tiny [u]}}$ we obtain $\nabla_ {\!a^{{\tiny [t]}}} \hat{J}_1(W, b) \text{{\Large $\vert$}} _{\hat{a}^{{\tiny [u]}}} \; = \; (a_1^{{\tiny [t]}} - \hat{a}_1^{{\tiny [u]}}, \; a_2^{{\tiny [t]}} - \hat{a}_2^{{\tiny [u]}}, \; \dots \: , \; a_{s_{n_l}}^{{\tiny [t]}} - \hat{a}_{s_{n_l}}^{{\tiny [u]}} )$. On the other hand, we have $\tilde{B}(a^{{\tiny [t]}} ; \; \hat{a}^{{\tiny [u]}} ) = || a^{{\tiny [t]}}\: - \hat{a}^{{\tiny [u]}} ||^2 \; = \; \sum_{i = 1}^{s_{n_l}} \dfrac{1}{2}(a_i^{{\tiny [t]}}\: - \: \hat{a}_i^{{\tiny [u]}})^2$, from which we may compute: $\nabla_ {\!a^{{\tiny [t]}}} \tilde{B}(a^{{\tiny [t]}} ; \; \hat{a}^{{\tiny [u]}} ) = (a_1^{{\tiny [t]}} - \hat{a}_1^{{\tiny [u]}}, \; a_2^{{\tiny [t]}} - \hat{a}_2^{{\tiny [u]}}, \; \dots \: , \; a_{s_{n_l}}^{{\tiny [t]}} - \hat{a}_{s_{n_l}}^{{\tiny [u]}} )$. Hence, $ \nabla_ {\!a^{{\tiny [t]}}} \hat{J}_1(W, b) \text{{\Large $\vert$}} _{\hat{a}^{{\tiny [u]}}} \:=\: \nabla_ {\!a^{{\tiny [t]}}} \tilde{B}(a^{{\tiny [t]}} ; \; \hat{a}^{{\tiny [u]}} )$, as required. 
\end{proof}

\noindent Before moving on, let us quickly \lq save\rq a key implication of the above result for future reference:
\begin{equation}
\text{{\Huge [}} \text{{\huge (}}\dfrac{\partial a^{{\tiny [t]}}}{\partial W_{ij}^{(l)} }\text{{\huge )}}^{\!T}\nabla_ {\!a^{{\tiny [t]}}} \hat{J}_1(W, b) \text{{\Huge ]}} \text{{\Huge $\vert$}} _{\hat{a}^{{\tiny [u]}}} \: = \: \dfrac{\partial \tilde{B}(a^{{\tiny [t]}} ; \; \hat{a}^{{\tiny [u]}} )} {\partial W^{(l)}_{ij}}
\end{equation}
At this juncture, we pause to take stock of what we have achieved so far. Essentially, what we have achieved up to this point is a means for computing the first of the two terms required for the computation of $\dfrac{\partial \hat{J}_1(W, b)}{\partial W_{ij}^{(l)} }$ as given in Equation 12. Hence, my natural next line of action is to consider how to compute the second of the two terms, which is the quantity $\text{{\huge (}}\dfrac{\partial a^{{\tiny [u]}}}{\partial W_{ij}^{(l)} }\text{{\huge )}}^{\!T} \nabla_ {\!a^{{\tiny [u]}}} \hat{J}_1(W, b)$. As it turns out, the approach to the computation of $\text{{\huge (}}\dfrac{\partial a^{{\tiny [u]}}}{\partial W_{ij}^{(l)} }\text{{\huge )}}^{\!T} \nabla_ {\!a^{{\tiny [u]}}} \hat{J}_1(W, b) \text{{\Huge $\vert$}} _{\hat{a}^{{\tiny [t]}}}$ is perfectly analogous to that of $\text{{\huge (}}\dfrac{\partial a^{{\tiny [t]}}}{\partial W_{ij}^{(l)} }\text{{\huge )}}^{\!T}\nabla_ {\!a^{{\tiny [t]}}} \hat{J}_1(W, b) \text{{\Large $\vert$}} _{\hat{a}^{{\tiny [u]}}}$, which have already dealt with, so I will not go into all the details. Only two things are required. First, one needs to take advantage of the fact that $(-1)^2 = 1$, which allows one to re-write $\hat{J}_1(W, b)$ in the form $\hat{J}_1(W, b) = \dfrac{1}{2}|| a^{{\tiny [u]}}\: - \: a^{{\tiny [t]}}||^2$. Second, one considers a back-propagatable function of the form $\tilde{B}(a^{{\tiny [u]}} ; \; \hat{a}^{{\tiny [t]}} )= \dfrac{1}{2}|| a^{{\tiny [u]}}\: - \: \hat{a}^{{\tiny [t]}}||^2$ (Compare this with the form $\tilde{B}(a^{{\tiny [t]}} ; \; \hat{a}^{{\tiny [u]}} ) = \dfrac{1}{2}|| a^{{\tiny [t]}}\: - \: \hat{a}^{{\tiny [u]}}||^2$ which we had used earlier on). Based on this, the current scenario becomes perfectly analogous to the scenario we had earlier dealt with earlier on, while trying to show the claim, $ \text{{\Huge [}} \text{{\huge (}}\dfrac{\partial a^{{\tiny [t]}}}{\partial W_{ij}^{(l)} }\text{{\huge )}}^{\!T}\nabla_ {\!a^{{\tiny [t]}}} \hat{J}_1(W, b) \text{{\Huge ]}} \text{{\Huge $\vert$}} _{\hat{a}^{{\tiny [u]}}} \: = \: \text{{\huge (}}\dfrac{\partial a^{{\tiny [t]}}}{\partial W_{ij}^{(l)} }\text{{\huge )}}^{\!T} \nabla_ {\!a^{{\tiny [t]}}} \tilde{B}(a^{{\tiny [t]}} ; \; \hat{a}^{{\tiny [u]}} )$. Hence, with the analogy, it should be clear that, for the present scenario, we must have: $ \text{{\Huge [}} \text{{\huge (}}\dfrac{\partial a^{{\tiny [u]}}}{\partial W_{ij}^{(l)} }\text{{\huge )}}^{\!T}\nabla_ {\!a^{{\tiny [u]}}} \hat{J}_1(W, b) \text{{\Huge ]}} \text{{\Huge $\vert$}} _{\hat{a}^{{\tiny [t]}}} \: = \: \text{{\huge (}}\dfrac{\partial a^{{\tiny [u]}}}{\partial W_{ij}^{(l)} }\text{{\huge )}}^{\!T} \nabla_ {\!a^{{\tiny [u]}}} \tilde{B}(a^{{\tiny [u]}} ; \; \hat{a}^{{\tiny [t]}} )$. Furthermore, for the present situation, we have an analog of Equation 19 as follows:
\begin{equation}
\text{{\Huge [}} \text{{\huge (}}\dfrac{\partial a^{{\tiny [u]}}}{\partial W_{ij}^{(l)} }\text{{\huge )}}^{\!T}\nabla_ {\!a^{{\tiny [u]}}} \hat{J}_1(W, b) \text{{\Huge ]}} \text{{\Huge $\vert$}} _{\hat{a}^{{\tiny [t]}}} \: = \: \dfrac{\partial \tilde{B}(a^{{\tiny [u]}} ; \; \hat{a}^{{\tiny [t]}} )} {\partial W^{(l)}_{ij}}
\end{equation}
\noindent Now, when we combine Equations 12, 19, and 20, it should not be hard to see that, for a given pair of constants $(\hat{a}^{{\tiny [t]}}, \hat{a}^{{\tiny [u]}})$, we must have:
\begin{equation}
\dfrac{\partial \hat{J}_1(W,b)}{\partial W_{ij}^{(l)} }\text{{\Huge $\vert$}}_{ (\hat{a}^{{\tiny [t]}}, \hat{a}^{{\tiny [u]}})} = \dfrac{\partial \tilde{B}(a^{{\tiny [t]}} ; \; \hat{a}^{{\tiny [u]}} )} {\partial W^{(l)}_{ij}}\text{{\Huge $\vert$}}_{ \hat{a}^{{\tiny [t]}}} \:+ \dfrac{\partial \tilde{B}(a^{{\tiny [u]}} ; \; \hat{a}^{{\tiny [t]}} )} {\partial W^{(l)}_{ij}} \text{{\Huge $\vert$}}_{ \hat{a}^{{\tiny [u]}}}
\end{equation}
A key essence of Equation 21 above is that it provides an avenue for computing $\dfrac{\partial \hat{J}_1(W,b)}{\partial W_{ij}^{(l)} }$,
for a given pair of constants $(\hat{a}^{{\tiny [t]}}, \hat{a}^{{\tiny [u]}})$, using the backpropagation algorithm. Let us illustrate the \lq avenue\rq $\:$ by considering how to compute the quantity $\dfrac{\partial \tilde{B}(a^{{\tiny [t]}} ; \; \hat{a}^{{\tiny [u]}} )} {\partial W^{(l)}_{ij}} \text{{\Huge $\vert$}}_{ \hat{a}^{{\tiny [t]}}}$, which occurs in the Equation. To compute $\dfrac{\partial \tilde{B}(a^{{\tiny [t]}} ; \; \hat{a}^{{\tiny [u]}} )} {\partial W^{(l)}_{ij}} \text{{\Huge $\vert$}}_{ \hat{a}^{{\tiny [t]}}}$, we simply call the standard backpropagation procedure, passing the current training example, $\hat{a}^{{\tiny [u]}}$ and $x^{{\tiny [t]}}$. In the procedure, $\hat{a}^{{\tiny [u]}}$ plays the role of the target output vector; while $x^{{\tiny [t]}}$ is used to calculate $\hat{a}^{{\tiny [t]}}$, which then plays the role of the current vector of output-layer activations from the ANN. On the flip side, to compute $\dfrac{\partial \tilde{B}(a^{{\tiny [u]}} ; \; \hat{a}^{{\tiny [t]}} )} {\partial W^{(l)}_{ij}} \text{{\Huge $\vert$}}_{ \hat{a}^{{\tiny [u]}}}$, we pass the training example, $x^{\tiny [u]}$ (not $x^{\tiny [t]}$), along with the pair, $\hat{a}^{{\tiny [t]}}$ (not $\hat{a}^{{\tiny [u]}}$), as arguments to the standard backpropagation procedure. This time, $\hat{a}^{{\tiny [u]}}$, which can be computed from $x^{\tiny [u]}$, plays the role of the current vector of activations produced by the ANN, while $\hat{a}^{{\tiny [t]}}$ plays the role of the target output vector. Based on the foregoing, we naturally derive the following algorithm for computing $\dfrac{\partial \hat{J}_1(W, b)}{\partial W_{ij}^{(l)} }$:
\begin{algorithm}[H]
\caption{ Harnessesing the Standard Back-Propagation (SBP) Procedure for Computing $\dfrac{\partial \hat{J}_1(W, b)}{\partial W_{ij}^{(l)} }$ }
\begin{algorithmic}
\STATE {\textbf{Inputs}: A set of current ANN weights, $W_{ij}^{(l)}$, and biases, $b_{i}^{(l)}$, $l = 1,2, \dots, n_l$, $i = 1, 2, \dots, s_{l}$, and $j = 1,2, \dots, s_{l-1}$, along with a pair of input vectors $(x^{{\tiny [t]}}, x^{{\tiny [u]}})$}.
\STATE
\STATE 1). Use the input vector, $x^{{\tiny [t]}}$, to perform an initialisation by filling ANN \lq input activations\rq \; with $ x^{{\tiny [t]}}$. 
\STATE \hspace{\algorithmicindent} $ \: $ That is, in the input layer, FOR each $i = 1, 2, \dots, s_{1}$, set:
\STATE \hspace{\algorithmicindent} \hspace{\algorithmicindent} $\quad \quad \;$ $ a_i^{(1)} := x_i^{{\tiny [t]}}$
\STATE \hspace{\algorithmicindent} $ \: $ END FOR
\STATE
\STATE 2). Perform a Feedforward pass through the ANN, computing all activations through out the
network. 
\STATE \hspace{\algorithmicindent} $ \: $ That is, FOR each $l = 1,2, \dots, n_l- 1$, FOR each $i = 1, 2, \dots, s_{l+1}$, FOR each $j = 1,2, \dots, s_{l}$, set:
\STATE \hspace{\algorithmicindent} \hspace{\algorithmicindent} 2a). $ \: $ $ z_i^{(l+1){\tiny [t]}} := W_{ij}^{(l)}a_i^{(l)} \: + \: b_i^{(l)}$
\STATE \hspace{\algorithmicindent} \hspace{\algorithmicindent} 2b). $ \: $ $a_i^{(l+1){\tiny [t]}} := f(z_i^{(l+1){\tiny [t]}})$
\STATE \hspace{\algorithmicindent} $ \: $ END FOR, $\:$ END FOR, $\:$ END FOR
\STATE
\STATE 3). Obtain $\hat{a}^{{\tiny [t]}}$ as the vector of output layer activations computed in STEP 2b above. 
\STATE \hspace{\algorithmicindent} $ \: $ That is, FOR each $i = 1, 2, \dots, s_{n_l}$, set: 
\STATE \hspace{\algorithmicindent} \hspace{\algorithmicindent} $\quad \quad \;$ $\hat{a}_i^{{\tiny [t]}} := a_i^{(n_l){\tiny [t]}}$.
\STATE \hspace{\algorithmicindent} $ \: $ END FOR
\STATE 
\STATE 4). Repeat STEPS 1 to 3, but this time using $x^{{\tiny [u]}}$, rather than $x^{{\tiny [t]}}$ (and using variable $u$ rather \STATE \hspace{\algorithmicindent} $ \: $ than $t$), thereby obtaining vector $\hat{a}^{{\tiny [u]}}$. 
\STATE
\STATE 5). Use the SBP procedure to compute $\dfrac{\partial \tilde{B}(a^{{\tiny [t]}} ; \; \hat{a}^{{\tiny [u]}} )} {\partial W^{(l)}_{ij}} \text{{\Huge $\vert$}}_{ \hat{a}^{{\tiny [t]}}}$, by passing $x^{{\tiny [t]}}$ and $\hat{a}^{{\tiny [u]}}$ to the 
\STATE \hspace{\algorithmicindent} $ \: $ procedure; this first time, $x^{{\tiny [t]}}$ plays the role of the current ANN input, while $\hat{a}^{{\tiny [u]}}$ plays the role 
\STATE \hspace{\algorithmicindent} $ \: $ of the target output vector: 
\STATE \hspace{\algorithmicindent} \hspace{\algorithmicindent} $\quad \quad \;$ $\Delta \tilde{B}_1 := \dfrac{\partial \tilde{B}(a^{{\tiny [t]}} ; \; \hat{a}^{{\tiny [u]}} )} {\partial W^{(l)}_{ij}} \text{{\Huge $\vert$}}_{ \hat{a}^{{\tiny [t]}}}$
\STATE
\STATE
\STATE 6). Use the SBP procedure to compute $\dfrac{\partial \tilde{B}(a^{{\tiny [u]}} ; \; \hat{a}^{{\tiny [t]}} )} {\partial W^{(l)}_{ij}} \text{{\Huge $\vert$}}_{ \hat{a}^{{\tiny [u]}}}$, by passing $x^{{\tiny [u]}}$ and $\hat{a}^{{\tiny [t]}}$ to the 
\STATE \hspace{\algorithmicindent} $ \: $ procedure; this second time, $x^{{\tiny [u]}}$ plays the role of the current ANN input, while $\hat{a}^{{\tiny [t]}}$ plays the role 
\STATE \hspace{\algorithmicindent} $ \: $ of the target output vector: 
\STATE \hspace{\algorithmicindent} \hspace{\algorithmicindent} $\quad \quad \;$ $\Delta \tilde{B}_2 := \dfrac{\partial \tilde{B}(a^{{\tiny [u]}} ; \; \hat{a}^{{\tiny [t]}} )} {\partial W^{(l)}_{ij}} \text{{\Huge $\vert$}}_{ \hat{a}^{{\tiny [u]}}}$
\STATE
\STATE
\STATE 7). Compute the final result:
\STATE \hspace{\algorithmicindent} \hspace{\algorithmicindent} $\quad \quad \;$ $\dfrac{\partial \hat{J}_1(W, b)}{\partial W_{ij}^{(l)} } := \Delta \tilde{B}_1 + \Delta\tilde{B}_2 $
\end{algorithmic} 
\end{algorithm}
\section{Deriving the Partial Derivatives of $\dfrac{\partial \hat{J}_2(W,b )}{\partial W_{ij}^{(l)} }$ and an Algorithm for Computing them.}
In the preceding section, we derived an expression, and corresponding algorithm, for computing $\dfrac{\partial \hat{J}_1(W,b )}{\partial W_{ij}^{(l)} }$. In this section, we derive an expression for $\dfrac{\partial \hat{J}_2(W,b )}{\partial W_{ij}^{(l)} }$ and furnish an algorithm for its computation. To proceed, let us recall from Section 3 that:
\begin{equation}
\hat{J}_2(W,b ) \:=\: || a^{(n_l)[t]} || _1
\end{equation}
Now, keeping the notation of Section 2 in mind, the above can also be written as:
\begin{equation}
\hat{J}_2(W,b ) = \sum_{i=1}^{s_{n_l}}|| a_i^{(n_l)[t]} || 
\end{equation}
In what follows, we will simply write $|| a_i^{(n_l)} || _1 $ instead of $|| a_i^{(n_l)[t]} || _1 $, for a less clumsy notation, and the reader is implored to keep this in mind. The ultimate plan is to derive a variant of the backpropagation algorithm for the computation of $\dfrac{\partial \hat{J}_2(W,b )}{\partial W_{ij}^{(l)} }$ for all $l$, $i$ and $j$. To this end, using the notation of Section 2, we first need to spell out the pertinent feed-forward equations:
\begin{equation}
z_i ^{(l+1)} = \sum_{j=1} ^{s_l} a_i^{(l)} w_{ij}^ {(l)} + b_i^{(l)}
\end{equation}
\begin{equation}
a_i^{(l)} = f(z_i^{(l)})
\end{equation}
where $f(.)$ denotes the relevant transfer function, which typically is the sigmoid, tan-sigmoid or linear transfer function, depending on the layer in question. Also, we define a \lq sign function \rq according to:
\begin{equation} 
sign(a_i) \; = \; 
\begin{cases} +1 & \text{if $a_i > 0$} \\ - 1 & \text{if $a_i < 0$} \\ 0 & \text{if $a_i = 0$}
\end{cases}
\end{equation} 
Based on the above, one can show, using the chain rule, that $\dfrac{\partial \hat{J}_2(W,b )}{\partial W_{ij}^{(n_l - 1)} }$, which is the partial derivative with respect to an arbitrary weight that projects to the output layer, $l = n_l$, from the penultimate layer, $l = n_l - 1$, is given by:
\begin{equation}
\dfrac{\partial \hat{J}_2(W,b )}{\partial W_{ij}^{(n_l - 1)} } \: = \: sign(a_i^{(n_l)})f'(z_i^{(n_l)})a_j^{(n_l - 1)}
\end{equation}
where $f'(.)$ denotes the derivative of the transfer function $f(.)$. In anticipation of what is to come next, we define a \lq signed derivative\rq, denoted $\beta_i^{(n_l)}$, as follows:
\begin{equation}
\beta_i^{(n_l)} \: = \: sign(a_i^{(n_l)})f'(z_i^{(n_l)})
\end{equation}
Let us notice the similarity between the form of $\beta_i^{(n_l)}$ and the so-called scaled error, $\delta_i^{(n_l)} = (a_i^{(n_l)} - y_i )f'(z_i^{(n_l)})$ (where $y_i$ denotes the $i$-th component of the current target output training vector), which plays a key role in the standard backpropagation algorithm. In particular, let us make the \emph{key observation} that $sign(a_i^{(n_l)})$ in $\beta_i^{(n_l)}$ is playing the role which $(a_i^{(n_l)} - y_i )$ plays in $\delta_i^{(n_l)}$. Indeed, we can easily express $\dfrac{\partial \hat{J}_2(W,b )}{\partial W_{ij}^{(n_l - 1)} }$ in terms of $\beta_i^{(n_l)}$ by putting Equation 28 into Equation 27:
\begin{equation}
\dfrac{\partial \hat{J}_2(W,b )}{\partial W_{ij}^{(n_l - 1)} } \: = \: \beta_i^{(n_l)} a_j^{(n_l - 1)}
\end{equation}
To proceed, let us now move one layer back through the ANN, and attempt to write an expression for $\dfrac{\partial \hat{J}_2(W,b )}{\partial W_{ij}^{(n_l - 2)} }$. Again, using the chain rule, one can show that: 
\begin{equation}
\dfrac{\partial \hat{J}_2(W,b )}{\partial W_{ij}^{(n_l - 2)} } \: = \: \sum_{p = 1} ^{s_{n_l}} sign(a_p^{(n_l)})f'(z_p^{(n_l)}) W_{pi}^{(n_l - 1)} f'(z_i^{(n_l - 1)}) a_j^{(n_l - 2)}
\end{equation}
But, from Equation 28, it is clear that for any index variable, $p$, we can write $\beta_p^{(n_l)} \: = \: sign(a_p^{(n_l)})f'(z_p^{(n_l)})$. Putting this into Equation 30, we readily obtain:
\begin{equation}
\dfrac{\partial \hat{J}_2(W,b )}{\partial W_{ij}^{(n_l - 2)} } \: = \: \sum_{p = 1} ^{s_{n_l}} \beta_p^{(n_l)} W_{pi}^{(n_l - 1)} f'(z_i^{(n_l - 1)}) a_j^{(n_l - 2)}
\end{equation}
At this juncture, we point out that if $\hat{J}_2(W,b )$ had been defined by $\hat{J}_2(W,b ) = \dfrac{1}{2}\sum_{i=1}^{s_{n_l}}( a_i^{(n_l)[t]} - y_i )^2$ rather than $\hat{J}_2(W,b ) = \sum_{i=1}^{s_{n_l}}|| a_i^{(n_l)[t]} ||$, then we would have had 
$\dfrac{\partial \hat{J}_2(W,b )}{\partial W_{ij}^{(n_l - 1)} } \: = \: \delta_i^{(n_l)} a_j^{(n_l - 1)}$ and $\dfrac{\partial \hat{J}_2(W,b )}{\partial W_{ij}^{(n_l - 2)} } \: = \: \sum_{p = 1} ^{s_{n_l}} \delta_p^{(n_l)} W_{pi}^{(n_l - 1)} f'(z_i^{(n_l - 1)}) a_j^{(n_l - 2)}$. This observation indicates that, for the problem at hand, we can obtain a backpropagation algorithm for computing the partial derivatives of $\hat{J}_2(W,b )$ simply by letting $\beta_i^{(l)}$ play the role which $\delta_i^{(l)}$ normally plays in the standard backpropagation algorithm. As a specific implication of the above observation, one could define $ \beta_i^{(n_l - 1)} \: = \: \sum_{p = 1} ^{s_{n_l}} \beta_p^{(n_l)} W_{pi}^{(n_l - 1)} f'(z_i^{(n_l - 1)})$, in analogy with what is usually done in the standard backpropagation algorithm setting, and then write $\dfrac{\partial \hat{J}_2(W,b )}{\partial W_{ij}^{(n_l - 2)} } \: = \: \beta_i^{(n_l - 1)} a_j^{(n_l - 2)}$. Now generalizing, for any $l \in \{ 1, 2, \dots, n_l - 1 \}$, we can define $\beta_i^{(l)}$ according to:
\begin{equation}
\beta_i^{(l)} \: = \: \sum_{p = 1} ^{s_{n_l}} \beta_p^{(l + 1)} W_{pi}^{(l)} f'(z_i^{(l)}) 
\end{equation}
where the computation of $\beta_i^{(n_l)}$ (i.e. the base step) is already given in Equation 31.
Hence, we are led to propose the algorithm below for computing the partial derivatives of $\hat{J}_2(W,b)$ with respect to ANN weights, (and biases as well):
\begin{algorithm}
\caption{A Back-Propagation Algorithm for Calculating $\dfrac{\partial \hat{J}_2(W,b)}{\partial W_{ij}^{(l)} }$ and $\dfrac{\partial \hat{J}_2(W,b)}{\partial b_{i}^{(l)} }$ }
\begin{algorithmic}
\STATE {\textbf{Inputs}: The set of current ANN weights, $W_{ij}^{(l)}$, and biases, $b_{i}^{(l)}$, $l = 1,2, \dots, n_l$, $i = 1, 2, \dots, s_{l}$, and $j = 1,2, \dots, s_{l-1}$, along with a specific input vector, $x^{{\tiny [t]}} = ( x_1^{{\tiny [t]}}, x_2^{{\tiny [t]}}, \dots, x_{s_1}^{{\tiny [t]}})$.}
\STATE
\STATE 1). Perform initialisation by filling ANN \lq input activations\rq \; with the input vector, $ x^{{\tiny [t]}}$. 
\STATE \hspace{\algorithmicindent}That is, in the input layer, FOR each $i = 1, 2, \dots, s_{1}$, set:
\STATE \hspace{\algorithmicindent} \hspace{\algorithmicindent} $ a_i^{(1)} := x_i^{{\tiny [t]}}$
\STATE \hspace{\algorithmicindent} END FOR
\STATE
\STATE 2). Perform a Feedforward pass through the ANN, computing all activations through out the
network. \STATE \hspace{\algorithmicindent}That is, FOR each $l = 1,2, \dots, n_l- 1$, for each $i = 1, 2, \dots, s_{l+1}$, for each $j = 1,2, \dots, s_{l}$, set:
\STATE \hspace{\algorithmicindent} \hspace{\algorithmicindent} 1a). $ z_i^{(l+1)} := W_{ij}^{(l)}a_i^{(l)} + b_i^{(l)}$
\STATE \hspace{\algorithmicindent} \hspace{\algorithmicindent} 1b). $a_i^{(l+1)} := f(z_i^{(l+1)})$
\STATE \hspace{\algorithmicindent} END FOR
\STATE
\STATE 3). At the output layer, (i.e. layer $L_{n_l}$), of the ANN, FOR each $i$ compute:
\STATE \hspace{\algorithmicindent} \hspace{\algorithmicindent} $\beta_i^{(n_l)} := sign(a_i^{(n_l)})f'(z_i^{(n_l)})$
\STATE \hspace{\algorithmicindent} END FOR
\STATE
\STATE 4). FOR each $l = 1,2, \dots, n_l- 1$, for each $i = 1, 2, \dots, s_{l+1}$, set:
\STATE \hspace{\algorithmicindent}\hspace{\algorithmicindent} 4a). $\dfrac{\partial \hat{J}_2(W,b )}{\partial W_{ij}^{(l)} } \: = \: \beta_i^{(l+1)} a_j^{(l)}$
\STATE \hspace{\algorithmicindent}\hspace{\algorithmicindent} 4b). $\dfrac{\partial \hat{J}_2(W,b )}{\partial b_{i}^{(l)} } \: = \: \beta_i^{(l+1)}$
\STATE \hspace{\algorithmicindent}\hspace{\algorithmicindent} 4c). $\beta_i^{(l)} \: = \: \sum_{j = 1} ^{s_{l+1}} \beta_j^{(l + 1)} W_{ji}^{(l)} f'(z_i^{(l)})$ 
\STATE \hspace{\algorithmicindent} END FOR
\end{algorithmic} 
\end{algorithm}
\section{A Gradient Descent Algorithm for Minimizing SENNS's Objective Function}
In this section, we describe a gradient descent algorithm for the minimisation of our objective function, which we gave in Equation 4. The algorithm is labeled as Algorithm 3 below. The algorithm relies on Algorithms 1 and 2 for computing $\dfrac{\partial \hat{J}_1(W,b)}{\partial W_{ij}^{(l)} }$ and $\dfrac{\partial \hat{J}_2(W,b)}{\partial W_{ij}^{(l)} }$ respectively. It then implicitly utilizes Equations 8, 9 of Section 3, for the computation of $\dfrac{\partial J_1(W,b)}{\partial W_{ij}^{(l)} }$ and $\dfrac{\partial J_2(W,b)}{\partial W_{ij}^{(l)} }$ respectively, before using Equation 7 of that same section to compute $\dfrac{\partial J(W,b)}{\partial W_{ij}^{(l)} }$. Finally, using a learning rate, $\alpha$, the algorithm proceeds to update ANN weights in a gradient-descent fashion according to:
\begin{equation}
W_{ij}^{(l)new} \:=\: W_{ij}^{(l)old} - \alpha \dfrac{\partial J(W,b)}{\partial W_{ij}^{(l)} }
\end{equation}
where $W_{ij}^{(l)old}$ and $W_{ij}^{(l)new}$ denote the old and new weights respectively. Without any further ado, here is the algorithm we set out to derive:
\begin{algorithm}[H]
\caption{ Minimisation of SENNS's Objective Function via Gradient Descent: Weights Version}
\begin{algorithmic}
\STATE {\textbf{Inputs}: Regularisation parameters, $\lambda_1, \lambda_2, \lambda_3, \lambda_4$, a learning rate, $\alpha$, a training set, $x^{{\tiny [t]}}, t = 1,2, \dots, m$, parameter $M_{\mathbb{C}}$, which is the number of training set pairs belonging to the same class, parameter $M_{\mathbb{D}}$, which is the number of training set pairs belonging to different classes, and a set of initial randomized small weights, $W_{ij}^{(l)}$, $l = 1,2, \dots, n_l$, $i = 1, 2, \dots, s_{l}$, and $j = 1,2, \dots, s_{l-1}$}.
\STATE
\STATE 1). Perform the pair of initialisations:
\STATE \hspace{\algorithmicindent} \hspace{\algorithmicindent} 1a). $ \: $ $\dfrac{\partial J_1(W,b)}{\partial W_{ij}^{(l)} } := 0$ 
\STATE \hspace{\algorithmicindent} \hspace{\algorithmicindent} 1b). $ \: $ $\Delta J_2 := 0$ 
\STATE
\STATE 2). FOR each pair $(x^{{\tiny [t]}}, x^{{\tiny [u]}}), t = 1,2, \dots, m, u = 1,2, \dots, m$
\STATE
\STATE \hspace{\algorithmicindent} \hspace{\algorithmicindent} 2a). $\:$ Using Equation 3, set $\mathbb{S}( x^{{\tiny [t]}}, x^{{\tiny [u]}} ) := \dfrac{\lambda_1}{M_ {\mathbb{C}}} $
\STATE \hspace{\algorithmicindent} \hspace{\algorithmicindent} 2b). $ \: $ Using the current pair, $(x^{{\tiny [t]}}, x^{{\tiny [u]}})$, compute $\dfrac{\partial \hat{J}_1(W, b)}{\partial W_{ij}^{(l)} }$ via Algorithm 1
\STATE \hspace{\algorithmicindent} \hspace{\algorithmicindent} 2c). $ \: $ Increment $ \dfrac{\partial J_1(W,b)}{\partial W_{ij}^{(l)} } := \dfrac{\partial J_1(W,b)}{\partial W_{ij}^{(l)} } \:+\: \mathbb{S}( x^{{\tiny [t]}}, x^{{\tiny [u]}} ) \dfrac{\partial \hat{J}_1(W, b)}{\partial W_{ij}^{(l)} }$
\STATE {\quad} END FOR
\STATE
\STATE 3). FOR each $x^{{\tiny [t]}}, t = 1,2, \dots, m$
\STATE \hspace{\algorithmicindent} \hspace{\algorithmicindent} 3a). $ \: $ Compute $\dfrac{\partial \hat{J}_2(W,b)}{\partial W_{ij}^{(l)} }$ via Algorithm 2
\STATE \hspace{\algorithmicindent} \hspace{\algorithmicindent} 3b). $ \: $ Increment $\Delta J_2 := \Delta J_2 + \dfrac{\partial \hat{J}_2(W,b)}{\partial W_{ij}^{(l)} }$
\STATE {\quad} END FOR
\STATE
\STATE 4). Set $\dfrac{\partial J_2(W,b)}{\partial W_{ij}^{(l)} } := \dfrac{\lambda_3}{m} \Delta J_2$
\STATE
\STATE 5). Set $\dfrac{\partial J(W,b)}{\partial W_{ij}^{(l)} } := \dfrac{\partial J_1(W,b)}{\partial W_{ij}^{(l)} } \:+ \: \dfrac{\lambda_3}{m} \dfrac{\partial J_2(W,b)}{\partial W_{ij}^{(l)} } 
\:+ \: \lambda_4 W_{ij}^{(l)}$
\STATE
\STATE 6). Finally, update the ANN weights via gradient descent using learning rate $\alpha$:
\STATE \hspace{\algorithmicindent} \hspace{\algorithmicindent} $W_{ij}^{(l)} = W_{ij}^{(l)} \: - \: \alpha{\Huge \text{(}} \dfrac{\partial J(W,b)}{\partial W_{ij}^{(l)} } {\Huge \text{)}}$
\STATE
\STATE 7). Repeat STEPs 1 to 6 untill convergence, or a maximum number of iterations is reached
\end{algorithmic} 
\end{algorithm}
\section{Conclusion}
In this concept paper, we have proposed a technique called SENNS (Sparse Extraction Neural NetworkS) for the feature extraction problem, which is a problem at the heart of pattern recognition and machine learning. Philosophically, our proposed method draws on the idea of extracting features that maximise inter-class variances, while minimising intra-class variances. As a result, the method fits immediately into the framework of graph embeddings. However, unlike two of the most representative members of the class of methods within the graph embedding school of thought, our proposed SENNS enforces sparsity on the extracted features, and utilises powerful non-linear projections rather than linear, kernel or tensor transformations, to effect the feature extraction process. We formulated SENNS as the minimisation of a regularised sum of four terms, and derived an effective gradient descent algorithm for the resulting minimisation problem. Via rigorous mathematical analysis, we showed how our algorithm can be specified as a set of tasks involving the standard back-propagation procedure, up to a modification for $L_1$ norms. Finally, a natural next line of action is to test SENNS out on some standard machine learning datasets such as the ARABASE database of Arabic characters, the CMU PIE database of faces, as well as the MNIST database of digits. 


\begin{thebibliography}{Abd1}
\bibitem{trier}
Trier, O.D., Jain, A.K. and Taxt, T. Feature Extraction Methods for Character Recognition: A Survey. \textit{Pattern Recognition}, vol 29, no 4, pp 641-662, 1996.
\bibitem{eig_map}
Belkin M. and Niyogi, P. Laplacian Eigenmaps and Spectral Techniques for Embedding and Clustering. \textit{Advances in Neural
Information Processing System}, vol. 14, pp. 585-591, 2001.
\bibitem{Lda}
Etemad K. and Chellapa, R. Discriminant Analysis for Recognition of Human Face Images. \textit {Journal of the Optical Society of America A}, vol. 14, no. 8 pp. 1724-1733, 1997.
\bibitem{turk_pent}
Turk, M. and Pentland, A. Face Recognition Using Eigenfaces, \textit{Proc. IEEE Conf. Computer Vision and Pattern Recognition}, pp. 586-591, 1991.
\bibitem{hu}
Hu, M.K. Visual Pattern Recognition by Moment Invariants. \textit{IRE Transactions on Information Theory}, vol. 8, pp 179-187, 1962.
\bibitem{ofmm}
Chao K. and Srinath M.D. Invariant Character Recognition with Zernike and Orthogonal Fourier Mellin Moments. \textit{Pattern Recognition} 35, 143-154, 2002.
\bibitem{sift}
Lowe, D. Distinctive Image Features from Scale-Invariant Keypoints. \textit {International Journal of Computer Vision} vol. 60, no. 2, pp. 91–110, 2004.
\bibitem{shape_cont}
Belongie, S., Malik J., and Puzicha J., Shape Matching and Object Recognition using Shape Contexts. \textit{IEEE Transactions on Pattern Analysis and Machine Intelligence}, 2002.
\bibitem{blum}
Blum, H. A. Transformation for Extracting New Descriptors of Shape. In: Models for the Perception of Speech and Visual Form, W. Wathen-Dunn, eds., \textit{MIT Press}, Cambridge, MA, pp. 362-380, 1967.
\bibitem{shock_graphs}
Kimia, B.B, Tannenbaum, A.R. and Zucker, S.W. Shape, Shocks and Deformations I: The Components of 2d Shape and the
Reaction-Diffusion Space, \textit {International Journal of Computer Vision} 15 189-224, 1995.
\bibitem{bone_graphs}
Macrini, D., Dickinson, S., Fleet, D., and Siddiqi, K. Bone Graphs: Medial Shape Parsing and Abstraction, \textit{Computer Vision and
Image Understanding} 115, 1044–1061, 2011.
\bibitem{yan_emb}
Yan, S., Xu, D., Zhang, B., Zhang, H.-J., Yang, Q. and Lin, S. Graph Embedding and Extensions: A General Framework for Dimensionality Reduction. \textit{ IEEE Transactionson Pattern Analysis and Machine Intelligence}, vol. 29, no. 1, pp. 40–51, 2007.
\bibitem{ganesh}
Sundaramoorthi, G. and Yang, Y. Matching Through Features and Features Through Matching. Technical Report, KAUST. 
\bibitem{dejiver}
Dejiver, P.A. and Kitler, J. \textit{Pattern Recognition: A Satistical Approach}. Prentice Hall, London, 1982.
\bibitem{arabase}
Ben Amara, N., Mazhoud, O., Bouzrara, N. and Ellouze, N. ARABASE: A Relational Database for Arabic OCR Systems. \textit{International Arab Journal of Information Technology}, 2005.
\bibitem{cmu_pie}
Sim, T., Baker, S., and Bsat, M. The CMU Pose, Illumination, and Expression Database. \textit{IEEE Trans. Pattern Analysis and Machine
Intelligence}, vol. 25, no. 12, pp. 1615-1618, Dec. 2003.
\bibitem{mnist}
LeCun, Y., Bottou, L., Bengio, Y. and Haffner, P. Gradient-based Learning Applied to Document Recognition. \textit{Proceedings of the IEEE}, 86(11):2278–2324, 1998.
\bibitem{Ng}
Ng A., Jiquan N., Chuan F., Yifan M. and Caroline S., \textit{UFDL Tutorial on Neural Networks}. Web. August 2014 $<$http://ufdl.stanford.edu/wiki/index.php/Neural\_Networks$>$
\bibitem{sparse_code}
Yang, J., Yu, K., Gong, Y. and Huang, T. Linear Spatial Pyramid Matching using Sparse Coding for Image Classification. In \textit{IEEE International Conference on Computer Vision and Pattern Recognition}, 2009.
\bibitem{lap_sparse}
Gao S., Tsang I., Chia L., and Zhao P. Local Features are not Lonely - Laplacian Sparse Coding for Image Classification. In \textit{IEEE International Conference on Computer Vision and Pattern Recognition}, 2010.
\end{thebibliography}
\end{document}